\documentclass{article}

\usepackage{PRIMEarxiv}

\usepackage[utf8]{inputenc} 
\usepackage[T1]{fontenc}    
\usepackage{hyperref}       
\usepackage{url}            
\usepackage{booktabs}       
\usepackage{amsfonts}       
\usepackage{nicefrac}       
\usepackage{microtype}      
\usepackage{lipsum}
\usepackage{fancyhdr}       
\usepackage{graphicx}       
\graphicspath{{media/}}     

\title{Artificial Intelligence in the Autonomous Navigation of Endovascular Interventions: A Systematic Review
\thanks{\textit{\underline{Citation}}: 
\textbf{Robertshaw H, Karstensen L, Jackson B, Sadati H, Rhode K, Ourselin S, Granados A and Booth TC (2023) Artificial intelligence in the autonomous navigation of endovascular interventions: a systematic review. \textit{Front. Hum. Neurosci.} 17:1239374. doi: 10.3389/fnhum.2023.1239374}} 
}

\author{
  Harry Robertshaw, Benjamin Jackson, Hadi Sadati, Kawal Rhode, Sebastien Ourselin, \\ \textbf{Alejandro Granados, Thomas C Booth} \\
  School of Biomedical Engineering \& Imaging Sciences \\
  Kings College London \\
  London\\
   \And
  Lennart Karstensen \\
  AIBE \\
  Friedrich-Alexander University Erlangen-Nürnberg \\
  Erlangen\\
}

\begin{document}
\maketitle

\begin{abstract}
    \textbf{Background} Autonomous navigation of catheters and guidewires in endovascular interventional surgery can decrease operation times, improve decision-making during surgery, and reduce operator radiation exposure while increasing access to treatment.
    
    \textbf{Objective} To determine from recent literature, through a systematic review, the impact, challenges, and opportunities artificial intelligence (AI) has for the autonomous navigation of catheters and guidewires for endovascular interventions.
    
    \textbf{Methods} PubMed and IEEEXplore databases were searched to identify reports of AI applied to autonomous navigation methods in endovascular interventional surgery. Eligibility criteria included studies investigating the use of AI in enabling the autonomous navigation of catheters/guidewires in endovascular interventions. Following Preferred Reporting Items for Systematic Reviews and Meta-Analysis (PRISMA), articles were assessed using Quality Assessment of Diagnostic Accuracy Studies 2 (QUADAS-2). PROSPERO: CRD42023392259.
    
    \textbf{Results} 462 studies fulfilled the search criteria, of which fourteen studies were included for analysis. Reinforcement learning (RL) (9/14, 64\%) and learning from expert demonstration (7/14, 50\%) were used as data-driven models for autonomous navigation. These studies evaluated models on physical phantoms (10/14, 71\%) and \textit{in-silico} (4/14, 29\%) models. Experiments within or around the blood vessels of the heart were reported by the majority of studies (10/14, 71\%), while non-anatomical vessel platforms `idealised' for simple navigation were used in three studies (3/14, 21\%), and the porcine liver venous system in one study. We observed that risk of bias and poor generalisability were present across studies. No procedures were performed on patients in any of the studies reviewed. Moreover, all studies were limited due to the lack of patient selection criteria, reference standards, and reproducibility, which resulted in a low level of evidence for clinical translation.
    
    \textbf{Conclusions} Despite the potential benefits of AI applied to autonomous navigation of endovascular interventions, the field is in an experimental proof-of-concept stage, with a technology readiness level of 3. We highlight that reference standards with well-identified performance metrics are crucial to allow for comparisons of data-driven algorithms proposed in the years to come.
\end{abstract}

\section{Introduction}

    Cardiovascular (CV) diseases are the most common cause of death across Europe, accounting for more than 4 million deaths each year, with coronary heart disease (44.2\%) and cerebrovascular disease (25.4\%) emerging as the predominant contributors to CV-related mortality across all ages and genders \cite{Townsend2016}. Endovascular catheter-based interventions such as percutaneous coronary intervention (PCI), pulmonary vein isolation (PVI) and mechanical thrombectomy (MT) have become an established treatment for CV diseases \cite{Thukkani2015, Goyal2016, Giacoppo2017, Lindgren2018}. During such a procedure, an operator navigates a guidewire and catheter from an insertion point (typically the common femoral or radial artery) to the area of interest to perform the intervention. Intraoperative fluoroscopy is used intermittently throughout the navigation and intervention to guide the catheter and guidewire through the vasculature. Once the target site has been reached, the treatment can be performed through the catheter. This is typically thrombus removal in the case of MT, stent deployment in the case of PCI, and ablation for PVI \cite{Brilakis2020}.

    In acute CV disease, time from symptom onset to treatment is often crucial for effective endovascular interventions. For example, the benefits of MT become non-significant after 7.3 hours of stroke for non-stratified patients \cite{Saver2016}. As a result, in the UK for example, only 1.4\% of stroke admissions benefit from MT despite the 10\% of patients that are eligible for treatment \cite{McMeekin2017}. Other challenges for endovascular interventions relate to occasional complications including perforation, thrombosis and dissection in the parent artery, as well as distal embolization of thrombus \cite{Hausegger2001}. Moreover, angiography requires intravascular contrast agent administration, which can occasionally lead to nephrotoxicity \cite{Rudnick1995}. For operators and their teams, the high cumulative dose of x-ray radiation from angiography is a risk factor for cancer and cataracts \cite{Klein2009}. Although exposure can be minimised with current radiation protection practice, some measures involve operators wearing heavy protective equipment which is a risk factor for orthopaedic complications, and so alternative methods of exposure reduction are beneficial \cite{Ho2007, Madder2017}.

    It is hoped that robotic surgical systems can either mitigate or eliminate some of the challenges currently presented by endovascular interventions. For example, robotic systems could be set up in hospitals nationwide and tele-operated remotely from a central location, increasing the speed of access to treatments such as MT beyond what is possible currently \cite{Crinnion2022}. Additionally, robotic systems might eliminate any operator physiological tremors or fatigue and allow endovascular interventions to be performed in an optimum ergonomic position while potentially increasing procedural precision (for example, procedure time), and thereby improving overall performance scores and reducing complication rates \cite{Riga2010}. Furthermore, as operators would not be required to stand next to the patient, their radiation exposure would be reduced and the need to wear heavy protective equipment would be obviated.

    Commercial robotic systems are currently available to perform endovascular interventions. Hansen Medical developed the \textit{Magellan}\textsuperscript{TM} system (Auris Health, Redwood City, USA), the first commercially available robotic system to be used for PVI, and more recently used to successfully perform carotid artery stenting in 13 patients \cite{Duran2014, Jones2021}. This system comprises a steerable guide catheter inside a steerable sheath allowing movement in three dimensions, and a separate remote guidewire manipulator allowing linear and rotational movement. The \textit{Corpath GRX}\textsuperscript{\textregistered} (Corindus Vascular Robotics, USA), the next-generation system of the \textit{Corpath}\textsuperscript{\textregistered} 200 robot, has successfully been used for PCI and PVI. This system has performed diagnostic cerebral angiography procedures and ten carotid artery stenting procedures \cite{Nogueira2020, Sajja2020, Weinberg2021}. Furthermore, it has been recently used to perform robot-assisted, neuroendovascular interventions including aneurysm embolisation and epistaxis embolisation \cite{Pereira2020, Cancelliere2022, Saber2022}. These systems use a controller-operator structure, where operators remotely control and navigate a robot through a patient's vasculature to the target site. In currently available systems, the operator has complete control over the robot and makes all of the decisions.

    While these robotic systems help alleviate some of the challenges of endovascular interventions, they have limitations. The controller-operator structure requires a reasonably high cognitive workload, can still result in human error and means that the procedure is limited to an individual operator's skill set \cite{Mofatteh2021}. These robotic systems also consist of user interfaces such as buttons and joysticks, requiring skills that are different to those used in current clinical practice. Additionally, a lack of haptic feedback from robotic systems might result in difficulties to receive tactile feedback from the catheters and guidewires as they interact with vessel walls \cite{Crinnion2022}.

    One emerging method of mitigating these challenges is using artificial intelligence (AI) techniques in conjunction with robotic systems. AI, and in particular, machine learning (ML), has accelerated in recent years in its applications for data analysis and learning \cite{Sarker2021}, with many areas of healthcare already making use of this technology for disease prediction and diagnosis \cite{Fatima2017, Silahtaroglu2021}. ML algorithms can be divided into three main groups: supervised, unsupervised, and reinforcement learning (RL). Supervised learning is the most common form of ML and involves constructing a model trained on a dataset with labels (the corresponding correct outputs). The model can then accurately predict the labels of new, unknown instances based on the patterns learned from the training data \cite{Kotsiantis2007}.

    Unsupervised learning involves training an algorithm to represent particular input features in a way that reflects the structure of the overall collection of input patterns \cite{Dayan2017}. In contrast to other types of ML, the dataset is unlabelled and there are no explicit target outputs or environmental evaluations associated with each input. 
    
    RL is a form of ML, whereby an agent learns by interacting with the environment and receiving feedback in the form of rewards. The goal of RL is to maximise the cumulative reward over time by learning a policy that optimises the agent's current state for a set of actions \cite{Arulkumaran2017}. Similar to the natural way of human learning, robotic RL automatically acquires the skills through `trials and errors' \cite{Sutton2018}. Applications of RL are becoming more expansive, as numerous research areas aim to use the method, for example, in precision medicine, medical imaging, and rehabilitation \cite{Lowery2013, Ghesu2018, Naros2015}. 

    Learning from demonstration (LfD) is a variant of supervised learning, where input data is provided by an expert demonstrator. This can also act as a precursor for RL, whereby the agent can further improve its behaviour through interaction with the environment. Table \ref{tab:ML} describes the ML methods that are referred to later in this paper, each of which can be used to improve performance across the three types of ML described above. LfD has been separated from the other types of ML in this case, as it can be used in the context of both supervised learning and RL. 

    \begin{table}[ht!]
        \centering
        \caption{ Description of ML methods.}
        \label{tab:ML}
        \renewcommand{\arraystretch}{1.25}
        \resizebox{\textwidth}{!}{%
        \begin{tabular}{lll}
        \hline
        \multicolumn{1}{|l|}{\textbf{Name}} &
          \multicolumn{1}{l|}{\textbf{ML Type}} &
          \textbf{Description} \\ \hline
        \multicolumn{1}{|l|}{A3C} &
          \multicolumn{1}{l|}{RL} &
          \multicolumn{1}{l|}{An algorithm that employs multiple agents working in parallel to learn policies in an environment. \cite{Mnih2016}.} \\ \hline
        \multicolumn{1}{|l|}{\begin{tabular}[c]{@{}l@{}}Behaviour \\ Cloning\end{tabular}} &
          \multicolumn{1}{l|}{LfD} &
          \multicolumn{1}{l|}{\begin{tabular}[c]{@{}l@{}}A technique where an agent learns a policy by imitating expert behaviour. It learns from labelled examples provided by \\ experts, mapping input observations to corresponding actions to replicate the demonstrated behaviour. It can be used as a \\ pre-training step in RL allowing the agent to learn by imitating the behaviour of an expert \cite{Codevilla2019}.\end{tabular}} \\ \hline
        \multicolumn{1}{|l|}{CNN} &
          \multicolumn{1}{l|}{\begin{tabular}[c]{@{}l@{}}Supervised \\ learning\end{tabular}} &
          \multicolumn{1}{l|}{\begin{tabular}[c]{@{}l@{}}Type of deep neural network specifically designed for image processing and pattern recognition tasks. CNNs leverage \\ spatial hierarchies through convolutional layers that extract local features and preserve spatial relationships, enabling \\ effective image classification, object detection, and image segmentation tasks \cite{OShea2015}.\end{tabular}} \\ \hline
        \multicolumn{1}{|l|}{DDPG} &
          \multicolumn{1}{l|}{RL} &
          \multicolumn{1}{l|}{\begin{tabular}[c]{@{}l@{}}An algorithm that merges RL and policy optimisation. It iteratively refines the policy based on estimated value distributions, \\ to find an optimal strategy. \cite{Lillicrap2015}.\end{tabular}} \\ \hline
        \multicolumn{1}{|l|}{DQN} &
          \multicolumn{1}{l|}{RL} &
          \multicolumn{1}{l|}{\begin{tabular}[c]{@{}l@{}}Leverages a deep neural network to learn optimal policies through Q-learning (see Q-learning explanation below). It \\ enables agents to make decisions by maximising the expected cumulative rewards, facilitating dynamic environment \\ interaction \cite{Mnih2013}.\end{tabular}} \\ \hline
        \multicolumn{1}{|l|}{Dueling DQN} &
          \multicolumn{1}{l|}{RL} &
          \multicolumn{1}{l|}{\begin{tabular}[c]{@{}l@{}}An extension of DQN that separates the estimation of state value and action advantages. By independently approximating \\ these values, the agent can learn the value of being in a particular state while also considering the advantages of each \\ action \cite{Wang2015}.\end{tabular}} \\ \hline
        \multicolumn{1}{|l|}{GAIL} &
          \multicolumn{1}{l|}{LfD} &
          \multicolumn{1}{l|}{\begin{tabular}[c]{@{}l@{}}Method where an agent learns a policy by imitating expert behaviour using a generative adversarial framework. It involves \\ a generator network that aims to replicate the expert and a discriminator network that distinguishes between expert and \\ generated behaviour \cite{Ho2016}.\end{tabular}} \\ \hline
        \multicolumn{1}{|l|}{GMM} &
          \multicolumn{1}{l|}{\begin{tabular}[c]{@{}l@{}}Unsupervised \\ learning\end{tabular}} &
          \multicolumn{1}{l|}{A statistical model that assumes data is generated by a mixture of several Gaussian distributions \cite{Reynolds2009}.} \\ \hline
        \multicolumn{1}{|l|}{\begin{tabular}[c]{@{}l@{}}HD\end{tabular}} &
          \multicolumn{1}{l|}{LfD} &
          \multicolumn{1}{l|}{\begin{tabular}[c]{@{}l@{}}Term that encompasses the process of an expert performing a task. Human demonstration can be used as a means to collect \\ data for LfD \cite{Nair2017}.\end{tabular}} \\ \hline
        \multicolumn{1}{|l|}{HER} &
          \multicolumn{1}{l|}{RL} &
          \multicolumn{1}{l|}{Allows an agent to learn from ``failed” experiences by redefining the goal of a task \cite{Andrychowicz2017}.} \\ \hline
        \multicolumn{1}{|l|}{HMM} &
          \multicolumn{1}{l|}{\begin{tabular}[c]{@{}l@{}}Unsupervised \\ learning\end{tabular}} &
          \multicolumn{1}{l|}{\begin{tabular}[c]{@{}l@{}}A statistical model that assumes observations are generated by a hidden sequence of states that follow a Markov process \\ \cite{Rabiner1989}.\end{tabular}} \\ \hline
        \multicolumn{1}{|l|}{PI\textsuperscript{2}} &
          \multicolumn{1}{l|}{RL} &
          \multicolumn{1}{l|}{\begin{tabular}[c]{@{}l@{}}Optimisation algorithm which aims to find the optimal policy by iteratively improving the policy through gradient-based \\ optimisation methods, maximising the expected return \cite{Theodorou2010}.\end{tabular}} \\ \hline
        \multicolumn{1}{|l|}{PPO} &
          \multicolumn{1}{l|}{RL} &
          \multicolumn{1}{l|}{\begin{tabular}[c]{@{}l@{}}An algorithm that optimises policies iteratively while ensuring small policy updates. It balances exploration and exploitation, \\ enhancing stability and sample efficiency during training \cite{Schulman2017}.\end{tabular}} \\ \hline
        \multicolumn{1}{|l|}{Q-learning} &
          \multicolumn{1}{l|}{RL} &
          \multicolumn{1}{l|}{\begin{tabular}[c]{@{}l@{}}Algorithm that learns the optimal action-value function (Q-value function) for sequential decision-making. It updates \\ Q-values iteratively based on observed rewards and the maximum expected future rewards \cite{Jang2019}.\end{tabular}} \\ \hline
        \multicolumn{1}{|l|}{Rainbow} &
          \multicolumn{1}{l|}{RL} &
          \multicolumn{1}{l|}{\begin{tabular}[c]{@{}l@{}} Extension of DQN that combines multiple improvements to enhance performance, by incorporating techniques such as \\ prioritised experience replay, distributional value estimation, and multi-step learning to improve overall learning stability \\ and efficiency \cite{Hessel2017}.\end{tabular}} \\ \hline
        \multicolumn{1}{|l|}{YOLO} &
          \multicolumn{1}{l|}{\begin{tabular}[c]{@{}l@{}}Supervised \\ learning\end{tabular}} &
          \multicolumn{1}{l|}{\begin{tabular}[c]{@{}l@{}}Object detection algorithm that can detect and classify objects in real-time. It uses a single neural network to directly \\ predict bounding boxes and class probabilities for objects in an image, providing fast and accurate object detection \\ \cite{Redmon2015}.\end{tabular}} \\ \hline
        \multicolumn{3}{l}{\begin{tabular}[c]{@{}l@{}}A3C, asynchronous advantage actor critic; CNN, convolutional neural network; DDPG, deep deterministic policy gradient; DQN, deep Q-network; GAIL, \\ generative adversarial imitation learning GMM, Gaussian mixture modelling; HD, human demonstration; HER, hindsight experience replay; HMM, hidden \\ Markov models; LfD, learning from demonstration; PI2, policy improvement with path integrals; PPO, proximal policy optimisation; RL, reinforcement \\ learning; YOLO, you only look once.\end{tabular}} \\
        \end{tabular}%
        }
    \end{table}

    The use of these ML techniques for autonomy in medical robotics presents several challenges. To help in the consideration of regulatory, ethical, and legal barriers imposed, a six-level autonomy framework has been proposed, ranging from no autonomy at level 0, up to level 5 which involves full autonomy with no human intervention \cite{Yang2017}. This study aims to systematically review the methodology, performance and autonomy level of AI applied to the autonomous navigation of catheters and guidewires for endovascular interventions. Understanding the current developments in the field will help determine the impact, challenges, and opportunities required to direct future translational research and ultimately guide clinical practice.

\section{Methods}

    This systematic review is PROSPERO (International prospective register of systematic reviews) registered (CRD42023392259). The review followed Preferred Reporting Items for Systematic Reviews and Meta-Analyses (PRISMA) guidelines \cite{Page2021}.

    \subsection{Selection criteria}

        \subsubsection{Eligibility criteria}
    
            Included reports consisted of primary research studies, which investigated the use of AI in enabling the autonomous navigation of catheters and/or guidewires in endovascular interventions. Excluded studies did not use AI methods to achieve autonomous navigation of catheters/guidewires or looked at path planning for endovascular interventions rather than the navigation itself. Additionally, studies without an English translation were not included \cite{Nussbaumer2020}.

        \subsubsection{Information sources and search strategy}

            PubMed and IEEEXplore were used to capture original research articles, published anytime until the end of January 2023, with the following search query: ``(Artificial Intelligence OR Machine Learning OR Reinforcement Learning OR Deep Learning OR Autonomous OR Learning-based) AND (Endovascular OR Vascular Intervention OR Catheter OR Guidewire) AND (Navigation OR Guidance)". Pre-prints and non-peer-reviewed articles were excluded.
    
        \subsubsection{Selection and data collection process}

            A medical robotics data scientist, H.R. (3 years of research experience), searched for studies as defined in the search strategy and followed the selection process as shown in Figure \ref{fig:PRISMA}. A medical robotics data scientist, L.K. (4 years experience in autonomous endovascular navigation using AI), independently reviewed the manuscripts against the eligibility criteria. In the case of discrepancy, consensus was reached by discussion between the two reviewers. If consensus was not reached, the multi-disciplinary authorship would make the final arbitration. The relevant data items, as defined in the following section, were extracted.

    \subsection{Data items, effect measures and synthesis methods}

        Information extracted from each study included: the AI method used and more granular model details (where available), the current level of autonomy, the type of experiment (\textit{in vivo}, \textit{in vitro}, \textit{in silico}), the method of tracking the catheter and/or guidewire position, the method of catheter and/or guidewire manipulation, description of the navigation path, performance measures, and key performance outcomes (where available).

        The levels of autonomy followed \cite{Yang2017}. Briefly, these are level 0: no autonomy, level 1: robot assistance, level 2: task autonomy, level 3: conditional autonomy, level 4: high autonomy, and level 5: full autonomy. It should be noted that if the autonomy level was not described in the study, an appropriate level was assigned based on the content of the paper.

    \subsection{Study risk of bias, reporting bias and certainty assessment}

        Where appropriate, both Quality Assessment of Diagnostic Accuracy Studies 2 (QUADAS-2) methodology alongside AI metrics from the Checklist for Artificial Intelligence in Medical Imaging (CLAIM) were used to assess the risk of bias for each study \cite{Rutjes2011, Mongan2020}.

\section{Results}

    \subsection{Studies}

\begin{figure}[ht!]
            \begin{center}
                \includegraphics[width=15cm]{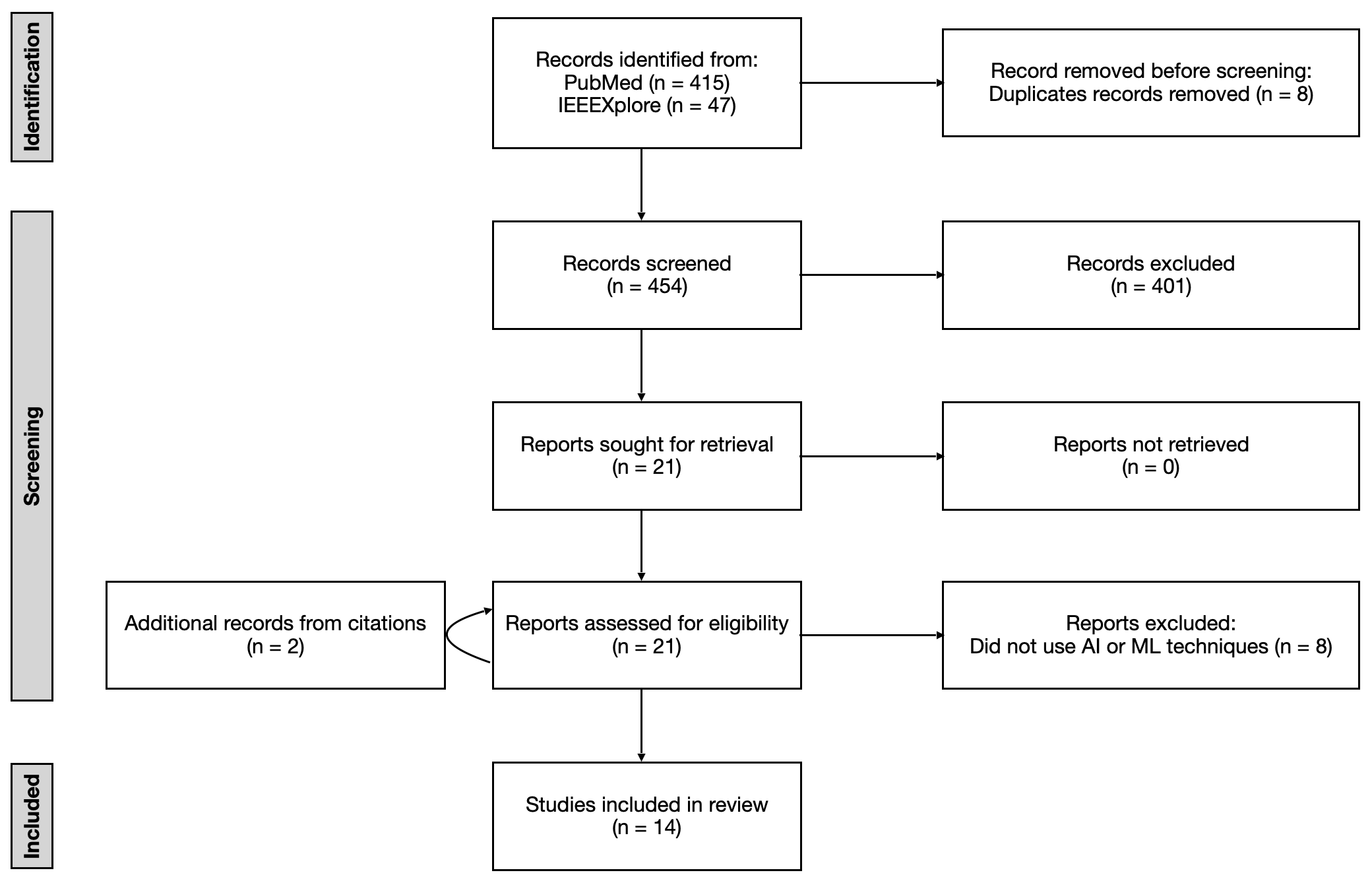}
            \end{center}
            \caption{ Preferred Reporting Items for Systematic Reviews and Meta-Analyses (PRISMA) flow diagram showing the number of articles searched and excluded at each stage of the literature search after screening titles, abstracts, and full texts.}\label{fig:PRISMA}
        \end{figure}

        As shown in Figure \ref{fig:PRISMA}, 462 studies met the search criteria, and 21 full-text studies were assessed against the eligibility criteria. A total of 14 were identified for review \cite{Chi2018_1, RafiiTari2013, Zhao2019, RafiiTari2014, Karstensen2022, Chi2018_2, Meng2022, Meng2021, Kweon2021, Cho2022, You2019, Behr2019, Wang2022, Chi2020}. The characteristics of the fourteen studies are listed in Table \ref{tab:Results}.

        \begin{table}[ht!]
            \centering
            \caption{ Studies resulting from our search and eligibility criteria proposing AI models for the autonomous navigation of catheters/guidewires in endovascular interventions. LfD was used as a ML method in cases where no further information about the type of LfD was available. Descriptions of each type of ML method can be found in Table \ref{tab:ML}.}
            \label{tab:Results}
            \renewcommand{\arraystretch}{1.2}
            \resizebox{\textwidth}{!}{%
            \begin{tabular}{|llllllll|}
            \hline
            \multicolumn{1}{|l|}{\textbf{Authors}} &
              \multicolumn{1}{l|}{\textbf{ML Method}} &
              \multicolumn{1}{l|}{\textbf{\begin{tabular}[c]{@{}l@{}}Level of \\ Autonomy\end{tabular}}} &
              \multicolumn{1}{l|}{\textbf{\begin{tabular}[c]{@{}l@{}}Validation \\ Environment\end{tabular}}} &
              \multicolumn{1}{l|}{\textbf{\begin{tabular}[c]{@{}l@{}}Tracking \\ Type\end{tabular}}} &
              \multicolumn{1}{l|}{\textbf{\begin{tabular}[c]{@{}l@{}}Tracking \\ Method\end{tabular}}} &
              \multicolumn{1}{l|}{\textbf{Navigation Path}} &
              \multicolumn{1}{l|}{\textbf{Performance Measures}} \\ \hline
              
            \multicolumn{8}{|l|}{\begin{tabular}[c]{@{}l@{}}\textbf{RL:}\end{tabular}} \\ \hline
            \multicolumn{1}{|l|}{\begin{tabular}[c]{@{}l@{}}Chi et al. \\ (2018b)\end{tabular}} &
              \multicolumn{1}{l|}{\begin{tabular}[c]{@{}l@{}}PI\textsuperscript{2} + LfD\end{tabular}} &
              \multicolumn{1}{l|}{Level 2} &
              \multicolumn{1}{l|}{\begin{tabular}[c]{@{}l@{}}\textit{In vitro} \\ (phantom)\end{tabular}} &
              \multicolumn{1}{l|}{\begin{tabular}[c]{@{}l@{}}Passive \\ (tracking-based)\end{tabular}} &
              \multicolumn{1}{l|}{\begin{tabular}[c]{@{}l@{}}EM position \\ sensor\end{tabular}} &
              \multicolumn{1}{l|}{\begin{tabular}[c]{@{}l@{}}Origin of LCA, to \\ BCA or LSA\end{tabular}} &
              \multicolumn{1}{l|}{\begin{tabular}[c]{@{}l@{}}Max acceleration of catheter tip,\\ Mean/max/STDEV/impact area of contact force,\\ Path-following error (RMSE),\\ Mean/STDEV speed of catheter tip,\\ Path length,\\ Procedure time\end{tabular}} \\ \hline
            \multicolumn{1}{|l|}{\begin{tabular}[c]{@{}l@{}}Behr et al. \\ (2019)\end{tabular}} &
              \multicolumn{1}{l|}{\begin{tabular}[c]{@{}l@{}} DDPG, DQN, HER, + HD\end{tabular}} &
              \multicolumn{1}{l|}{Level 3} &
              \multicolumn{1}{l|}{\begin{tabular}[c]{@{}l@{}}\textit{In vitro} \\ (phantom)\end{tabular}} &
              \multicolumn{1}{l|}{\begin{tabular}[c]{@{}l@{}}Passive (image-\\ based)\end{tabular}} &
              \multicolumn{1}{l|}{\begin{tabular}[c]{@{}l@{}}Top-down \\ camera\end{tabular}} &
              \multicolumn{1}{l|}{\begin{tabular}[c]{@{}l@{}}Idealised vessel \\ platform with a \\ bifurcation followed \\ by a bi-and \\ trifurcation in one \\ plane\end{tabular}} &
              \multicolumn{1}{l|}{SR of navigation task} \\ \hline
            \multicolumn{1}{|l|}{\begin{tabular}[c]{@{}l@{}}You et al. \\ (2019)\end{tabular}} &
              \multicolumn{1}{l|}{\begin{tabular}[c]{@{}l@{}}Dueling DQN\end{tabular}} &
              \multicolumn{1}{l|}{Level 3} &
              \multicolumn{1}{l|}{\begin{tabular}[c]{@{}l@{}}\textit{In vitro} \\ (phantom)\end{tabular}} &
              \multicolumn{1}{l|}{\begin{tabular}[c]{@{}l@{}}Passive \\ (tracking-based)\end{tabular}} &
              \multicolumn{1}{l|}{\begin{tabular}[c]{@{}l@{}}EM position \\ sensor\end{tabular}} &
              \multicolumn{1}{l|}{\begin{tabular}[c]{@{}l@{}}Insertion at heart \\ RA’s IVC, target is \\ nerve nodes around \\ CS and TA of heart \\ RA\end{tabular}} &
              \multicolumn{1}{l|}{\begin{tabular}[c]{@{}l@{}}Path length, \\ SR of navigation task\end{tabular}} \\ \hline
            \multicolumn{1}{|l|}{\begin{tabular}[c]{@{}l@{}}Chi et al. \\ (2020)\end{tabular}} &
              \multicolumn{1}{l|}{\begin{tabular}[c]{@{}l@{}}PPO + GAIL\end{tabular}} &
              \multicolumn{1}{l|}{Level 2} &
              \multicolumn{1}{l|}{\begin{tabular}[c]{@{}l@{}}\textit{In vitro} \\ (phantom)\end{tabular}} &
              \multicolumn{1}{l|}{\begin{tabular}[c]{@{}l@{}}Passive \\ (tracking-based)\end{tabular}} &
              \multicolumn{1}{l|}{\begin{tabular}[c]{@{}l@{}}EM position \\ sensor\end{tabular}} &
              \multicolumn{1}{l|}{\begin{tabular}[c]{@{}l@{}}Position in aorta \\ (proximal to major \\ branches), to BCA \\ or LCCA\end{tabular}} &
              \multicolumn{1}{l|}{\begin{tabular}[c]{@{}l@{}}Mean/max force between endovascular instruments \\ and vascular phantom,\\ Mean/STDEV speed of catheter tip,\\ Path length,\\ Procedure time, \\SR of navigation task\end{tabular}} \\ \hline
            \multicolumn{1}{|l|}{\begin{tabular}[c]{@{}l@{}}Cho et al. \\ (2022)\end{tabular}} &
              \multicolumn{1}{l|}{\begin{tabular}[c]{@{}l@{}}DDPG + Behaviour Cloning\end{tabular}} &
              \multicolumn{1}{l|}{Level 3} &
              \multicolumn{1}{l|}{\begin{tabular}[c]{@{}l@{}}\textit{In vitro} \\ (phantom)\end{tabular}} &
              \multicolumn{1}{l|}{\begin{tabular}[c]{@{}l@{}}Passive (image-\\ based)\end{tabular}} &
              \multicolumn{1}{l|}{\begin{tabular}[c]{@{}l@{}}Top-down \\ camera\end{tabular}} &
              \multicolumn{1}{l|}{\begin{tabular}[c]{@{}l@{}}Idealised vessel \\ platform with a \\ bifurcation followed \\ by a bifurcation in one \\ plane\end{tabular}} &
              \multicolumn{1}{l|}{Procedure time} \\ \hline
            \multicolumn{1}{|l|}{\begin{tabular}[c]{@{}l@{}}Meng et al. \\ (2021)\end{tabular}} &
              \multicolumn{1}{l|}{A3C} &
              \multicolumn{1}{l|}{Level 1} &
              \multicolumn{1}{l|}{\textit{In silico}} &
              \multicolumn{1}{l|}{\begin{tabular}[c]{@{}l@{}}Passive \\ (tracking-based)\end{tabular}} &
              \multicolumn{1}{l|}{\begin{tabular}[c]{@{}l@{}}Simulation-\\ based\end{tabular}} &
              \multicolumn{1}{l|}{\begin{tabular}[c]{@{}l@{}}Traversing descending \\ aorta, through aortic \\ arch, cannulation of \\ LCA, LSA or \\ innominate artery\end{tabular}} &
              \multicolumn{1}{l|}{\begin{tabular}[c]{@{}l@{}}Limited information available\end{tabular}} \\ \hline
            \multicolumn{1}{|l|}{\begin{tabular}[c]{@{}l@{}}Kweon et \\ al. (2021)\end{tabular}} &
              \multicolumn{1}{l|}{\begin{tabular}[c]{@{}l@{}}Rainbow + HD\end{tabular}} &
              \multicolumn{1}{l|}{Level 3} &
              \multicolumn{1}{l|}{\begin{tabular}[c]{@{}l@{}}\textit{In vitro} \\ (phantom)\end{tabular}} &
              \multicolumn{1}{l|}{\begin{tabular}[c]{@{}l@{}}Passive (image-\\ based)\end{tabular}} &
              \multicolumn{1}{l|}{\begin{tabular}[c]{@{}l@{}}Top-down \\ camera\end{tabular}} &
              \multicolumn{1}{l|}{\begin{tabular}[c]{@{}l@{}} Proximal point in left \\ anterior descending \\ artery to target \\ location in main or \\ side branch
              \end{tabular}} &
              \multicolumn{1}{l|}{\begin{tabular}[c]{@{}l@{}}Procedure time, \\SR of navigation task\end{tabular}} \\ \hline
            \multicolumn{1}{|l|}{\begin{tabular}[c]{@{}l@{}}Meng et al. \\ (2022)\end{tabular}} &
              \multicolumn{1}{l|}{A3C} &
              \multicolumn{1}{l|}{Level 1} &
              \multicolumn{1}{l|}{\textit{In silico}} &
              \multicolumn{1}{l|}{\begin{tabular}[c]{@{}l@{}}Passive \\ (tracking-based)\end{tabular}} &
              \multicolumn{1}{l|}{\begin{tabular}[c]{@{}l@{}}Simulation-\\ based\end{tabular}} &
              \multicolumn{1}{l|}{\begin{tabular}[c]{@{}l@{}}Traversing descending \\ aorta, through aortic \\ arch, cannulation of \\ LCA, LSA or \\ innominate artery\end{tabular}} &
              \multicolumn{1}{l|}{\begin{tabular}[c]{@{}l@{}}Contact force, \\Procedure time\end{tabular}} \\ \hline
            \multicolumn{1}{|l|}{\begin{tabular}[c]{@{}l@{}}Karstensen \\ et al. (2022)\end{tabular}} &
              \multicolumn{1}{l|}{\begin{tabular}[c]{@{}l@{}}DDPG, HER\end{tabular}} &
              \multicolumn{1}{l|}{Level 3} &
              \multicolumn{1}{l|}{\begin{tabular}[c]{@{}l@{}}\textit{Ex vivo} \\ (porcine liver)\end{tabular}} &
              \multicolumn{1}{l|}{\begin{tabular}[c]{@{}l@{}}Passive (image-\\ based)\end{tabular}} &
              \multicolumn{1}{l|}{Fluoroscopy} &
              \multicolumn{1}{l|}{\begin{tabular}[c]{@{}l@{}}Vena cava inferior \\ to vena hepatica \\ dextra, vena hepatica \\ intermedia or and \\ vena hepatica sinistra \\ (porcine liver)\end{tabular}} &
              \multicolumn{1}{l|}{\begin{tabular}[c]{@{}l@{}}Number of failures due to wrong branch/entanglement, \\ SR of navigation task\end{tabular}} \\ \hline
              
            \multicolumn{8}{|l|}{\begin{tabular}[c]{@{}l@{}}\textbf{non-RL:}\end{tabular}} \\ \hline
            \multicolumn{1}{|l|}{\begin{tabular}[c]{@{}l@{}}Rafii-Tari et \\ al. (2013)\end{tabular}} &
              \multicolumn{1}{l|}{GMM + LfD} &
              \multicolumn{1}{l|}{Level 2} &
              \multicolumn{1}{l|}{\begin{tabular}[c]{@{}l@{}}\textit{In vitro} \\ (phantom)\end{tabular}} &
              \multicolumn{1}{l|}{\begin{tabular}[c]{@{}l@{}}Passive \\ (tracking-based)\end{tabular}} &
              \multicolumn{1}{l|}{\begin{tabular}[c]{@{}l@{}}EM position \\ sensor\end{tabular}} &
              \multicolumn{1}{l|}{\begin{tabular}[c]{@{}l@{}}Traversing descending \\ aorta, through aortic \\ arch, cannulation of \\ innominate artery\end{tabular}} &
              \multicolumn{1}{l|}{\begin{tabular}[c]{@{}l@{}}Mean/max acceleration of catheter tip,\\ Mean/max speed of catheter tip\end{tabular}} \\ \hline
            \multicolumn{1}{|l|}{\begin{tabular}[c]{@{}l@{}}Rafii-Tari et \\ al. (2014)\end{tabular}} &
              \multicolumn{1}{l|}{HMM + LfD} &
              \multicolumn{1}{l|}{Level 2} &
              \multicolumn{1}{l|}{\begin{tabular}[c]{@{}l@{}}\textit{In vitro} \\ (phantom)\end{tabular}} &
              \multicolumn{1}{l|}{\begin{tabular}[c]{@{}l@{}}Passive \\ (tracking-based)\end{tabular}} &
              \multicolumn{1}{l|}{\begin{tabular}[c]{@{}l@{}}EM position \\ sensor\end{tabular}} &
              \multicolumn{1}{l|}{\begin{tabular}[c]{@{}l@{}}Cannulation of LSA \\ and RCCA\end{tabular}} &
              \multicolumn{1}{l|}{\begin{tabular}[c]{@{}l@{}}Mean/max acceleration of catheter tip,\\ Path length\end{tabular}} \\ \hline
            \multicolumn{1}{|l|}{\begin{tabular}[c]{@{}l@{}}Chi et al. \\ (2018a)\end{tabular}} &
              \multicolumn{1}{l|}{GMM + LfD} &
              \multicolumn{1}{l|}{Level 2} &
              \multicolumn{1}{l|}{\begin{tabular}[c]{@{}l@{}}\textit{In vitro} \\ (phantom)\end{tabular}} &
              \multicolumn{1}{l|}{\begin{tabular}[c]{@{}l@{}}Passive \\ (tracking-based)\end{tabular}} &
              \multicolumn{1}{l|}{\begin{tabular}[c]{@{}l@{}}EM position \\ sensor\end{tabular}} &
              \multicolumn{1}{l|}{\begin{tabular}[c]{@{}l@{}}Origin of LCA, to \\ bifurcation site \\ between RCCA and \\ RSA\end{tabular}} &
              \multicolumn{1}{l|}{\begin{tabular}[c]{@{}l@{}}Mean/max acceleration of catheter tip,\\ Mean/max/STDEV/impact area of contact force,\\ Mean/max/STDEV speed of catheter tip,\\ Path length,\\ SR of navigation task\end{tabular}} \\ \hline
            \multicolumn{1}{|l|}{\begin{tabular}[c]{@{}l@{}}Zhao et al. \\ (2019)\end{tabular}} &
              \multicolumn{1}{l|}{\begin{tabular}[c]{@{}l@{}}CNN\end{tabular}} &
              \multicolumn{1}{l|}{Level 3} &
              \multicolumn{1}{l|}{\begin{tabular}[c]{@{}l@{}}\textit{In vitro} \\ (phantom)\end{tabular}} &
              \multicolumn{1}{l|}{\begin{tabular}[c]{@{}l@{}}Passive (image-\\ based)\end{tabular}} &
              \multicolumn{1}{l|}{\begin{tabular}[c]{@{}l@{}}Top-down \\ camera\end{tabular}} &
              \multicolumn{1}{l|}{\begin{tabular}[c]{@{}l@{}}Medical and \\ designed vessel \\ models\end{tabular}} &
              \multicolumn{1}{l|}{\begin{tabular}[c]{@{}l@{}}Procedure time, \\ SR of navigation task\end{tabular}} \\ \hline
            \multicolumn{1}{|l|}{\begin{tabular}[c]{@{}l@{}}Wang et al. \\ (2022)\end{tabular}} &
              \multicolumn{1}{l|}{YOLOV5s} &
              \multicolumn{1}{l|}{Level 3} &
              \multicolumn{1}{l|}{\begin{tabular}[c]{@{}l@{}}\textit{In vitro} \\ (phantom)\end{tabular}} &
              \multicolumn{1}{l|}{\begin{tabular}[c]{@{}l@{}}Passive (image-\\ based)\end{tabular}} &
              \multicolumn{1}{l|}{\begin{tabular}[c]{@{}l@{}}Top-down \\ camera\end{tabular}} &
              \multicolumn{1}{l|}{\begin{tabular}[c]{@{}l@{}}Femoral to coronary \\ artery\end{tabular}} &
              \multicolumn{1}{l|}{\begin{tabular}[c]{@{}l@{}}Average Precision\end{tabular}} \\ \hline
            \multicolumn{8}{l}{\begin{tabular}[c]{@{}l@{}}\textbf{Clinical}: BCA, brachiocephalic artery; CS, coronary sinus; EM, electromagnetic; IVC, inferior vena cava; LCA, left coronary artery; LCCA, left common \\ carotid artery; LSA, left subclavian artery; RA, right atrium; RCCA, right common carotid artery; RSA, right subclavian artery; TA, transaortic.\\ \textbf{Technical}: A3C, asynchronous advantage actor critic; CNN, convolutional neural network; DDPG, deep deterministic policy gradient; DQN, deep Q-network; GAIL, generative adversarial \\ imitation learning GMM, Gaussian mixture modelling; HD, human demonstration; HER, hindsight experience replay; HMM, hidden Markov models; LfD, learning from demonstration; PI\textsuperscript{2}, \\ policy improvement with path integrals; PPO, proximal policy optimisation; RL, reinforcement learning.\\ \textbf{Evaluation}: RMSE, root-mean-squared error; SR, success rate; STDEV, standard deviation.\end{tabular}} \\
            \end{tabular}%
            }
        \end{table}

    According to QUADAS-2 methodology, all studies reviewed gave a high or unclear `risk of bias' and `concerns regarding applicability' in all domains. No studies performed procedures on patients and therefore had no clearly defined patient selection criteria, reference standards, or index tests. Despite the low level of evidence, there is value in discussing these individual studies as they represent the current state of the art and form a baseline for further research.

    \subsection{AI models}

        \subsubsection{RL methods}

            RL was used in nine studies (9/14, 64\%) with algorithms including A3C, DDPG, DQN, Dueling DQN, HER, PI\textsuperscript{2}, PPO, and Rainbow \cite{Behr2019, Chi2018_2, Chi2020, Karstensen2022, Kweon2021, Meng2021, Meng2022, You2019, Cho2022}. Demonstrator data in some form (GAIL, Behaviour Cloning or HD) was used as a precursor in four of the studies (4/14, 29\%) during training (LfD), in conjunction with other RL algorithms \cite{Chi2018_2, Behr2019, Cho2022, Kweon2021}. The SOFA framework (Inria, Strasbourg, France) \cite{Faure2012} was used for training RL models in four studies (4/14, 29\%) \cite{Behr2019, Cho2022, Karstensen2022, Meng2022}, the Unity engine (Unity Technologies, San Francisco, USA) was used in two studies (2/14, 14\%) \cite{You2019, Meng2021}, while the platform used for training was not specified in three studies (3/14, 21\%) \cite{Chi2018_2, Kweon2021, Chi2020}.

        \subsubsection{Other ML types}

            RL was not used in five studies (5/14, 36\%) which employed LfD (not as a precursor for RL), unsupervised (GMM and HMM) and supervised (CNN and YOLO) methods alone or in combination \cite{RafiiTari2013, RafiiTari2014, Chi2018_1, Zhao2019, Wang2022}. The most common method was LfD, which was used in three studies (3/14, 21\%) \cite{RafiiTari2013, RafiiTari2014, Chi2018_1}. Two of these studies (2/14, 14\%) used a GMM to generate the probabilistic representation of the dataset provided by a demonstrator \cite{RafiiTari2013, Chi2018_1}, while the other study utilised HMMs to model each movement primitive \cite{RafiiTari2014}. The other two non-RL studies (2/14, 14\%) used solely CNNs or YOLOV5s \cite{Zhao2019, Wang2022}.
        
    \subsection{Level of autonomy}

        Conditional autonomy (level 3) was performed in seven studies (7/14, 50\%) \cite{Behr2019, You2019, Cho2022, Kweon2021, Karstensen2022, Zhao2019, Wang2022}. Here, a target in the vasculature is selected by an operator and the subsequent navigation to the target of the guidewire and/or catheter takes place autonomously. Task autonomy (level 2) was performed across five studies (5/14, 36\%), whereby the robotic driver automates the catheter motion and an operator manipulates the guidewire for assistance \cite{Chi2018_2, Chi2020, RafiiTari2013, RafiiTari2014, Chi2018_1}. Robot assistance (level 1) was demonstrated in two studies (2/14, 14\%), where experiments were performed entirely in simulation and under continuous supervision of an operator \cite{Meng2021, Meng2022}. 
        
    \subsection{Experimental design}

        Clinical trials were not performed in any of the studies reviewed. Physical phantoms were used in the majority of studies (11/14, 79\%) reviewed \cite{Chi2018_2, Behr2019, You2019, Chi2020, Cho2022, Kweon2021, RafiiTari2013, RafiiTari2014, Chi2018_1, Zhao2019, Wang2022}. Of these studies, seven used 3D vascular phantoms \cite{Chi2018_1, Chi2020, Chi2018_2, RafiiTari2013, RafiiTari2014, You2019, Wang2022}, three used 2D phantoms \cite{Behr2019, Cho2022, Zhao2019}, and one study used both 2D and 3D phantoms \cite{Kweon2021}. Commercial phantoms were used in six studies (6/14, 43\%): 3D silicone-based, transparent, anthropomorphic phantoms (Elastrat Sarl, Geneva, Switzerland) were used in 5/14 (36\%) studies, \cite{Chi2018_1, Chi2020, Chi2018_2, RafiiTari2013, RafiiTari2014}; and the study using both 2D and 3D phantoms used firstly, a 2D PCI trainer for beginners (Medi Alpha Co., Ltd., Tokyo, Japan) and secondly, a silicone 3D Embedded Coronary Model (Trandomed 3D Medical Technology Co., Ltd., Ningbo, China), respectively. Five studies (5/14, 36\%) appeared to use in-house phantoms: one study used a silicone-based 3D printed heart model and inferior vena cava \cite{You2019}, and one used a 10 mm vessel diameter phantom made of polymethyl methacrylate (PMMA) \cite{Behr2019}, while sufficient phantom detail is not provided by the other three studies \cite{Wang2022, Cho2022, Zhao2019}.

        \textit{In silico} methods were used in two of the studies (2/14, 14\%) (one used SOFA framework and one used Unity engine) \cite{Meng2021, Meng2022}. \textit{Ex vivo} experiments using porcine liver vasculature were reported by one study \cite{Karstensen2022}. Here, \textit{in silico} methods were used for training models before the \textit{ex vivo} experiments.

\begin{figure}[!h]
            \begin{center}
                \includegraphics[width=9cm]{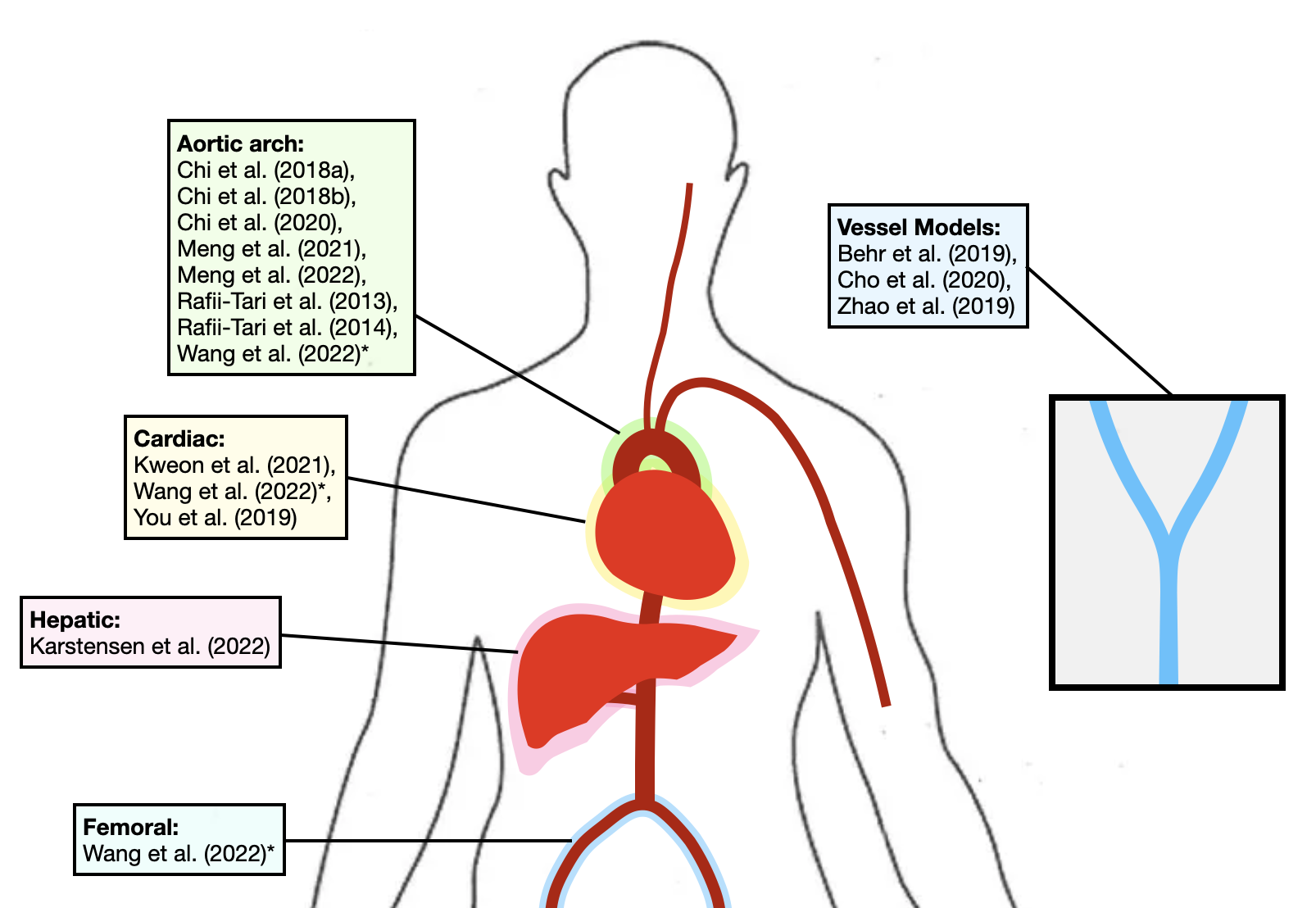}
            \end{center}
            \caption{ Diagram depicting the general vessels of interest for each study. *Study is in more than one area. Studies using non-anatomical platforms are also shown.}\label{fig:Vessel}
        \end{figure}

        Figure \ref{fig:Vessel} shows the anatomical regions where each study focuses. Experiments within or around the blood vessels of the heart were reported by the majority of studies (10/14, 71\%) \cite{Chi2018_1, Chi2018_2, Chi2020, Meng2021, Meng2022, RafiiTari2013, RafiiTari2014, Wang2022, Kweon2021, You2019}, with the study with the longest path length starting at the femoral artery and finishing at the coronary artery \cite{Wang2022}. Non-anatomical vessel platforms `idealised' for simple navigation were used in three studies (3/14, 21\%) \cite{Behr2019, Cho2022, Zhao2019}, and the porcine liver venous system in one study \cite{Karstensen2022}.

    \subsection{Evaluation}

        Passive tracking relies on external sensors to detect the catheter's position, active tracking involves the use of sensors located at the distal end of the catheter for real-time position tracking, and magnetic tracking utilizes external magnetic fields to guide the catheter's movement and track its position. A passive, tracking-based, method for catheter manipulation was used in eight studies (8/14, 57\%) \cite{Chi2018_1, Chi2018_2, You2019, Chi2020, Meng2021, Meng2022, RafiiTari2013, RafiiTari2014}, whereas a passive, image-based, method for catheter manipulation was used in the other six studies (6/14, 43\%) \cite{Behr2019, Cho2022, Kweon2021, Karstensen2022, Zhao2019, Wang2022}. None of the studies reviewed reported active or magnetic steering methods.
        
        A top-down camera for tracking the location of the guidewire and/or catheter was implemented in five of the studies (5/14, 36\%) where transparent phantoms allowed real-time video to provide software-generated tracking data \cite{Behr2019, Cho2022, Kweon2021, Zhao2019, Wang2022}. Electromagnetic (EM) position sensors were employed in six studies (6/14, 43\%) \cite{Chi2018_2, You2019, Chi2020, RafiiTari2013, RafiiTari2014, Chi2018_1}. An  Aurora control unit and EM Generator of Aurora electromagnetic tracking system (NDI, Waterloo, Canada) were used in one of these studies \cite{You2019}, whilst custom-designed sensors \cite{RafiiTari2013} were used in the other five. These five studies also employed a top-down camera simultaneously enabled through the use of transparent phantoms during data collection pre-training. One study employed continuous fluoroscopy, capturing 7.5 images per second, and used a CNN to segment the guidewire from real-time fluoroscopy images to track data that included the coordinates \cite{Karstensen2022}. Two studies (2/14, 14\%) were performed entirely \textit{in silico}, and hence no tracking method was required \cite{Meng2021, Meng2022}.

        Quantitative performance measures used in the studies were heterogeneous which may reflect the low technology readiness level (TRL) \cite{Mankins1995} of AI applied to autonomous navigation of endovascular interventions shown by the studies in this systematic review. Common performance measures used were success rate of navigation task (7/14, 50\%) and time to complete procedure (5/14, 36\%). Other performance measures shared across studies were: measures of force (6/14, 43\%); acceleration (4/14, 29\%); various measures of speed (4/14, 29\%); and path length (4/14, 29\%). Half of the studies (7/14, 50\%) reviewed compared manual performance against their autonomous navigation performance. The key performance outcomes of the fourteen studies are listed in Table \ref{tab:Outcomes}.

        \begin{table}[]
            \centering
            \caption{ Key performance outcomes from studies reviewed.}
            \label{tab:Outcomes}
            \renewcommand{\arraystretch}{1.25}
            \resizebox{\textwidth}{!}{%
            \begin{tabular}{|llll|}
            \hline
            \multicolumn{1}{|l|}{\textbf{Authors}} &
              \multicolumn{1}{l|}{\textbf{Path Length}} &
              \multicolumn{1}{l|}{\textbf{Procedure Time}} &
              \textbf{Success Rate} \\ \hline
            \multicolumn{1}{|l|}{\textbf{Behr et al. (2019)}} &
              \multicolumn{1}{l|}{n/a} &
              \multicolumn{1}{l|}{n/a} &
              70\% for DDPG \\ \hline
            \multicolumn{1}{|l|}{\textbf{Chi et al. (2018a)}} &
              \multicolumn{1}{l|}{\begin{tabular}[c]{@{}l@{}}Median: 360.5 mm manual, \\ 281.2 mm robot\end{tabular}} &
              \multicolumn{1}{l|}{n/a} &
              \begin{tabular}[c]{@{}l@{}}Expert model: 100\% under dry condition, \\ 94.4\% under continuous flow, \\ 55.6\% under pulsatile flow\end{tabular} \\ \hline
            \multicolumn{1}{|l|}{\textbf{Chi et al. (2018b)}} &
              \multicolumn{1}{l|}{\begin{tabular}[c]{@{}l@{}}367.8 mm pre-RL,\\ 211.6 mm RL\end{tabular}} &
              \multicolumn{1}{l|}{\begin{tabular}[c]{@{}l@{}}74.5 ± 11.6 s manual, \\ 137.7 ± 7.1 s pre-RL,\\ 121.3 ± 9.5 s RL\end{tabular}} &
              n/a \\ \hline
            \multicolumn{1}{|l|}{\textbf{Chi et al. (2020)}} &
              \multicolumn{1}{l|}{\begin{tabular}[c]{@{}l@{}}Type-I Aortic Arch, BCA: 55.7 ± 9.4 mm \\ automation, 51.4 ± 8.3 mm manual\end{tabular}} &
              \multicolumn{1}{l|}{\begin{tabular}[c]{@{}l@{}}Type-I Aortic Arch, BCA: 52.1 ± 9.9 s \\ automation, 6.36 ± 1.4 s manual\end{tabular}} &
              \begin{tabular}[c]{@{}l@{}}Type-I Aortic Arch: 94.4\% for BCA \\ cannulation, 88.9\% for LCCA cannulation\end{tabular} \\ \hline
            \multicolumn{1}{|l|}{\textbf{Cho et al. (2020)}} &
              \multicolumn{1}{l|}{n/a} &
              \multicolumn{1}{l|}{\begin{tabular}[c]{@{}l@{}}Real vessel phantom: 34.06 s  own \\ algorithm, 63.2 s expert algorithm\end{tabular}} &
              n/a \\ \hline
            \multicolumn{1}{|l|}{\textbf{Karstensen et al. (2022)}} &
              \multicolumn{1}{l|}{n/a} &
              \multicolumn{1}{l|}{n/a} &
              30\% (ex-vivo surgical task) \\ \hline
            \multicolumn{1}{|l|}{\textbf{Kweon et al. (2021)}} &
              \multicolumn{1}{l|}{n/a} &
              \multicolumn{1}{l|}{\begin{tabular}[c]{@{}l@{}}Proximal targets: 9.29 ± 6.00 s \\ autonomous, 82.1 ± 34.2 s manual\end{tabular}} &
              \textgreater{}95\% after 646 episodes (distal-main target) \\ \hline
            \multicolumn{1}{|l|}{\textbf{Meng et al. (2021)}} &
              \multicolumn{1}{l|}{n/a} &
              \multicolumn{1}{l|}{n/a} &
              n/a \\ \hline
            \multicolumn{1}{|l|}{\textbf{Meng et al. (2022)}} &
              \multicolumn{1}{l|}{n/a} &
              \multicolumn{1}{l|}{\begin{tabular}[c]{@{}l@{}}97.35 s manual,\\ 68.61 s training\end{tabular}} &
              n/a \\ \hline
            \multicolumn{1}{|l|}{\textbf{Rafii-Tari et al. (2013)}} &
              \multicolumn{1}{l|}{n/a} &
              \multicolumn{1}{l|}{n/a} &
              n/a \\ \hline
            \multicolumn{1}{|l|}{\textbf{Rafii-Tari et al. (2014)}} &
              \multicolumn{1}{l|}{\begin{tabular}[c]{@{}l@{}}2.9 m LSA manual intermediate,\\ 0.44 m LSA robot intermediate\end{tabular}} &
              \multicolumn{1}{l|}{n/a} &
              n/a \\ \hline
            \multicolumn{1}{|l|}{\textbf{Wang et al. (2022)}} &
              \multicolumn{1}{l|}{n/a} &
              \multicolumn{1}{l|}{n/a} &
              n/a \\ \hline
            \multicolumn{1}{|l|}{\textbf{You et al. (2019)}} &
              \multicolumn{1}{l|}{n/a} &
              \multicolumn{1}{l|}{n/a} &
              73\% no noise model (phantom) \\ \hline
            \multicolumn{1}{|l|}{\textbf{Zhao et al. (2019)}} &
              \multicolumn{1}{l|}{n/a} &
              \multicolumn{1}{l|}{n/a} &
              \begin{tabular}[c]{@{}l@{}}Medical vessel model: 94\%, \\ Designed vessel model: 92\%\end{tabular} \\ \hline
            \multicolumn{4}{l}{\begin{tabular}[c]{@{}l@{}}BCA, brachiocephalic artery; DDPG, deep deterministic policy gradient; LCCA, left common carotid artery; LSA, left subclavian artery; RL, reinforcement \\ learning.\end{tabular}} \\
            \end{tabular}%
            }
        \end{table}
        
        Where possible, critical outcome data for success rate, procedure time and path length were extracted from the study. Three of the fourteen studies (3/14, 21\%) did not measure any of these performance measures \cite{RafiiTari2013, Meng2021, Wang2022}. Of the seven studies (7/14, 50\%) that measured success rate, the value was over 90\% in four studies (4/14, 29\%) \cite{Chi2018_2, Chi2020, Kweon2021, Zhao2019}.

\section{Discussion}

    \subsection{Summary of findings}
    
        There is no high-level evidence \cite{Howick2011} to demonstrate that AI autonomous navigation of catheters and guidewires in endovascular intervention is non-inferior or superior to manual procedures. Currently, AI autonomous navigation of catheters and guidewires in endovascular intervention has not surpassed TRL 3. There has been no clinical validation nor has there been comprehensive laboratory validation. Over half of the studies (9/14, 64\%) employed RL methodologies, particularly in recent years, where most studies used RL (8/10, 80\% published beyond 2018). There are no standardised in silico, in vitro or ex vivo experimental reference standard designs, nor are there standardised performance measures, meaning comparison of studies quantitatively is of limited value.

    \subsection{Strengths and limitations}

        \subsubsection{Strengths}

            The primary strength of the studies reviewed came from the range of ML techniques employed. Most focused on finding a ML technique that would improve upon previous work, rather than using similar algorithms and extending the experimental environment. This is demonstrated well within the nine studies (9/14, 64\%) which used RL, where a different ML-based methodology was used in every case except for two (where the simulation environment and output measurements were changed between studies). Exploring various techniques is advantageous for research, especially in the rapidly evolving field of ML, as the fast pace of development increases the likelihood that more effective algorithms are created. For example, autonomous endovascular intervention progress has been catalysed by combining two recent approaches (LfD and RL) \cite{Chi2018_2, Chi2020, Cho2022, Kweon2021}. Here, using demonstrator data in a third of the RL studies allowed expert operator skill in complex endovascular procedures to be incorporated. This proficiency can be leveraged effectively to accelerate the RL training process. The combined approach, therefore, shortens the transition from a simulated training environment to a physical testing environment which typically presents significant challenges, as evidenced by the findings of \cite{Karstensen2022}. Another benefit of accelerating the process is that in some scenarios thousands of mechanical experimental training cycles may no longer be required leading to reduced mechanical wear on the experimental equipment.

        \subsubsection{Limitations}

            The limitations of the studies assessed encompassed three areas:

            \paragraph{}
            
                Whilst it was a strength that most studies focused on finding a ML technique that would improve upon previous endovascular navigation, the lack of focus on using similar or fixed algorithms and extending the experimental environment was a limitation. The challenge of fixing many experimental variables whilst changing another, is compounded by the lack of standardised \textit{in silico}, \textit{in vitro} or \textit{ex vivo} experimental reference standard designs for endovascular navigation, as well as a lack of standardised performance measures. As such, the ability to compare studies quantitatively was limited by confounding. For example, although some performance measures (e.g., `success rate’ and `procedure time’) were common to several studies, study comparison was limited due to experimental variations between studies. Firstly, the navigation path used to test the models varied. Secondly, some studies defined `success rate’ only if a task was completed within a certain time frame, whereas others had no time limit for completion. Thirdly, `procedure time’ was measured using different starting points and target sites.

            \paragraph{}

                Another limitation, also concerned with reference standards, is the importance of comparing the endovascular navigation with an autonomous system against the endovascular navigation without an autonomous system, to determine any incremental benefit through autonomy. Critically, the endovascular navigation without an autonomous system should ideally be operated by a relevant expert operating with minimal technical constraint to derive the reference standard (baseline) allowing comparison. Half the studies (7/14, 50\%) reviewed did compare endovascular navigation with and without an autonomous system; however, in some cases, the operator was technically constrained by using a novel robotic system rather than using the equipment used and processes they would typically employ, during an endovascular procedure in the clinic. For example, the robotic systems used in the studies reviewed failed to mimic the haptic feedback that the operator would receive performing procedures manually, such as viscous forces between the catheters and the blood; friction forces between the catheters and the vessel wall, and impact forces from the tips of the catheter and guidewire, and the vessel wall \cite{Crinnion2022}. Additionally, an expert is not able to use their previous experience with standard equipment and may be unfamiliar with these controls, meaning that performance at a given task will likely be affected.

            \paragraph{}

                There were no clinical studies of autonomous endovascular navigation which is a reflection of the nascent field and current TRL of the technology. The majority of studies (11/14, 79\%) were \textit{in vitro} and are valuable for development and testing as they limit the number of failures during subsequent in vivo testing \cite{Ionita2014}. However, these studies did not consider whether construct, face, and context validity of endovascular navigation systems was acceptable to allow TRL progression towards the clinic. In particular, in many of the studies reviewed, there were translational concerns regarding how the guidewires and/or catheters are tracked within the vasculature, as the alternative to using fluoroscopy with standard off-the-shelf catheters and guidewires is to create entirely new tracking methods. For example, several papers (6/14, 43\%) used EM-tracking to visualise the catheter in real-time, which has been shown to allow better real-time 3D orientation, facilitating navigation, reducing cannulation and total fluoroscopy times, and improving motion consistency and efficiency \cite{Schwein2017}. However, clinical translation using this method would require the introduction of new systems with specialised catheters and guidewires, resulting in additional costs and training.  Furthermore, other studies (5/14, 36\%) employed an experimental set-up involving a tabletop with a transparent phantom and a top-down camera. In its current state, this tracking method would not be suitable for future clinical studies, as a top-down camera would not be able to provide images of the guidewire and/or catheter through patient tissue. Nonetheless, it is noted that top-down cameras have a narrower clinical translation gap than EM-tracking, as they pose the same 2D challenges as fluoroscopy. 




    \subsection{Final thoughts and future research}

        Using AI, it may be possible to create a robotic system capable of autonomously navigating catheters and wires through a patient’s vasculature to the target site, requiring minimal assistance from an operator. If proven to be safe and effective in clinical trials, the benefits of autonomous navigation are numerous. It is plausible that in clinical specialities facing a shortage of highly-trained operators, there may be a reduced need for their expertise, potentially leading to greater accessibility of endovascular treatments globally, such as MT. For example, components of MT such as complex navigation tasks could be performed autonomously. Furthermore, autonomous systems are not limited by human factors such as fatigue or loss of focus, potentially making procedures safer and quicker \cite{Mirnezami2018}.

        The concept of fully autonomous navigation in endovascular interventions is promising; however, with a TRL level of 3 \cite{Mankins1995}, the technology is yet to complete validation even in a laboratory environment. Due to the inadequate evidence supporting its use (the limited number of studies and its low-level) \cite{Howick2011}, it is far from being used in clinical practice. It first must be demonstrated that it can reliably provide benefits over currently available treatments before it can progress towards clinical trials.

        Importantly, reference standards for endovascular navigation models need to be established to allow new models to be compared. This would allow effective comparison of different AI methods to determine the most effective model for autonomous endovascular navigation. These reference standards need to be established judiciously at the \textit{in silico}, \textit{in vitro}, and \textit{ex vivo} level with carefully-defined environments for different endovascular tasks such as PCI, PVI, and MT. It is noteworthy that at the \textit{in silico} level, where there are continuous advancements in modelling research and increased computational power, other areas of clinically-orientated ML research have successfully employed reference standards to enable reproducibility of results and comparability between competing models \cite{Russakovsky2015, Stubbs2019}. This includes computer vision (ImageNet Large Scale Visual Recognition Challenge) and natural language processing (National NLP Clinical Challenges). Furthermore, a set of minimum reporting standards of performance should be defined for studies investigating the use of AI in the autonomous navigation of endovascular interventions. In combination with a reference standard, this would allow complete comparison between ML algorithms designed for this specific task. 

        Clear regulation is required to determine how the community designs systems for the autonomous navigation in endovascular interventions. In the seven studies (7/14, 50\%) which proposed a system with `level 3’ autonomy, there is an expert operator in place who can intervene in the autonomous task if needed (`human in the loop’). It may be prudent, for now, for researchers to focus on optimising systems with `level 1-3’ autonomy as at higher levels of autonomy, particularly `level 5’ and potentially `level 4’, it is unclear how systems, where the robot can make decisions, will be regulated. As such future researchers may wish to optimise simple task autonomy, for example the autonomous navigation from the puncture point to the target site, in a system where an operator can stop the procedure and take over at any time. It is envisaged that as autonomous technology and regulations mature over time, systems will then be updated to carry out more difficult tasks.


        Various AI methods have been used to investigate the possibility of autonomous navigation in endovascular interventions. Although it is plausible that autonomous navigation may eventually benefit patients while reducing occupational hazards for staff, there is currently no high-level evidence to support this assertion. For the technology to progress, reference standards and minimum reporting standards need to be established to allow meaningful comparisons of new system development.

\section*{Conflict of Interest Statement}
The authors declare that the research was conducted in the absence of any commercial or financial relationships that could be construed as a potential conflict of interest.

\section*{Author Contributions}
H.R., A.G.M., and T.C.B.  contributed to the study conception and design. H.R. carried out the paper identification and screening. L.K. reviewed the manuscripts selected by H.R. against the eligibility criteria. The first draft of the manuscript was written by H.R. and edited by A.G.M., and T.C.B. All authors commented on versions of the manuscript. All authors read and approved the manuscript.

\section*{Funding}
Partial financial support was received from the WELLCOME TRUST under Grant Agreement No 203148/A/16/Z and from the Engineering and Physical Sciences Research Council Doctoral Training Partnership (EPSRC DTP) under Grant Agreement No EP/R513064/1.

\bibliographystyle{unsrt}  
\bibliography{references}

\end{document}